\documentclass[10pt,twocolumn,letterpaper]{article}

\usepackage{iccv}
\usepackage{times}
\usepackage{epsfig}
\usepackage{graphicx}
\usepackage{amsmath}
\usepackage{amssymb}

\usepackage{subfigure}
\usepackage{color}
\usepackage{url}
\usepackage{float}

\usepackage[pagebackref=true,breaklinks=true,letterpaper=true,colorlinks,bookmarks=false]{hyperref}

\iccvfinalcopy 

\def\iccvPaperID{****} 

\ificcvfinal\fi

\begin{document}

\title{Towards Flops-constrained Face Recognition}
\newcommand*{\affaddr}[1]{#1} 
\newcommand*{\affmark}[1][*]{\textsuperscript{#1}}
\newcommand*{\email}[1]{\texttt{#1}}
\author{%
Yu Liu\thanks{They contributed equally to this work}\affmark[ \ 1]
 \ \ \ \ \ \ \ \ Guanglu Song\footnotemark[1]\affmark[ \ 2]
 \ \ \ \ \ \ \ \ Manyuan Zhang\footnotemark[1]\affmark[ \ 2]
 \ \ \ \ \ \ \ \ Jihao Liu\footnotemark[1]\affmark[ \ 2]\\
 Yucong Zhou\affmark[2] \ \ \ \ \ \ \ \ Junjie Yan\affmark[2]\\
\texttt{yuliu@ee.cuhk.edu.hk} \\
\affaddr{\affmark[1]The Chinese University of Hong Kong}\\
\texttt{\{songguanglu,zhangmanyuan,liujihao\}@sensetime.com}
 \\
\affaddr{\affmark[2]SenseTime Research}
}

\maketitle
\ificcvfinal\thispagestyle{empty}\fi


\begin{abstract}
    Large scale face recognition is challenging especially when the computational budget is limited. Given a \textit{flops} upper bound, the key is to find the optimal neural network architecture and optimization method. In this article, we briefly introduce the solutions of team 'trojans' for the ICCV19 - Lightweight Face Recognition Challenge~\cite{lfr}. The challenge requires each submission to be one single model with computational budget no higher than 30 GFlops. We introduce a searched network architecture `Efficient PolyFace' based on the Flops constraint, a novel loss function `ArcNegFace', a novel frame aggregation method `QAN++', together with a bag of useful tricks in our implementation (augmentations, regular face, label smoothing, anchor finetuning, etc.). Our basic model, `Efficient PolyFace', takes 28.25 Gflops for the `deepglint-large' image-based track, and the `PolyFace+QAN++' solution takes 24.12 Gflops for the `iQiyi-large' video-based track. These two solutions achieve 94.198\% @ 1e-8 and 72.981\% @ 1e-4 in the two tracks respectively, which are the state-of-the-art results\footnote{The 72.981\% result wins the \textbf{1st} place on the IQIYI-large track and the 94.198\% wins the 2nd place on deepglint-large. However, the result on deepglint-large needs further deliberation. Note that our 94.189\% result on deepglint-large is adjusted by AdaBN, which uses image-level information of test set. For a fair comparison, the accuracy of Efficient PolyFace w/o AdaBN is 93.801\% as shown in Tab.~\ref{tab:adabn}} in this competition.
\end{abstract}

\section{Lightweight Face Recognition Challenge}
The ICCV19-Lightweight Face Recognition Challenge~\cite{lfr} is one of the most strict competitions in open-set face recognition. It requires the strict consistency of training data~\cite{guo2016ms}, face detector~\cite{deng2019retinaface} and alignment method between different submissions. There are four tracks in this competition: small image-based, large image-based, small video-based and large video-based. The computational budged is 1Gflops and 30Gflops for the small and large tracks respectively.

\section{Image-based baseline model}
We adopt two different CNN architectures R100~\cite{Deng_2019_CVPR} and a proposed PolyFace as our base models.
The input sizes of the two basic architectures are both $112\times112$ as required by the challenge~\cite{lfr}. 

\begin{figure*}
\centering
\includegraphics[width=1\linewidth]{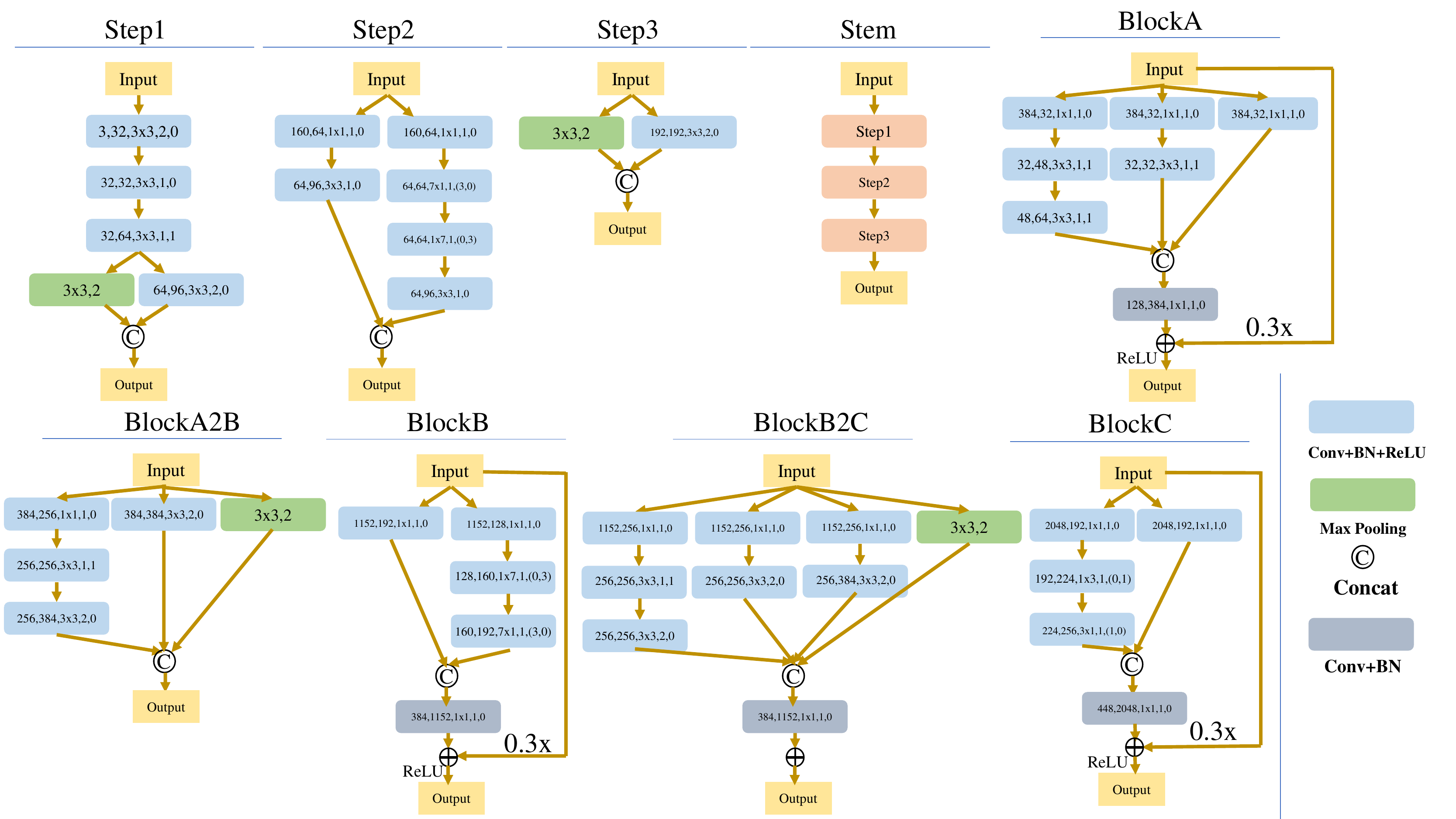}
   \caption{The details of blocks in PolyFace.
   The numbers in block $Conv+BN+ReLU$ represent the input channel, output channel, kernel size, stride, and padding. The numbers in block $Max$ $Pooling$ represent the kernel size and stride. The numbers in block $Conv+BN$ represent the input channel, output channel, kernel size, stride, and padding.}
\label{fig:apoly}
\end{figure*}

\textbf{PolyFace.} Similar to the structure of PolyNet~\cite{zhang2017polynet}, the basic PolyFace is designed by repeating its basic blocks. Details of the basic blocks are shown in Fig~\ref{fig:apoly}. In the stem block of the proposed PolyFace, the spatial size is first upsampled to $235 \times 235$ and then downsized to $112\times112$ by an upsampling and a convolutional layer, which we call 'stem-enrichment block'.
The data flow in the whole PolyFace is:

\texttt{Stem block -- A $\times$ blockA -- blockA2B -- B$\times$ blockB -- blockB2C -- C$\times$ blockC}. 

At the end of all backbones, a fully connected layer with 256 out-channels is adopted to generate the representation, followed by a \texttt{BatchNorm1d} layer. The block number of [A,B,C] in base model is [10,20,10].

\textbf{Training details.} During the training process of the base models, 16 GPUs are used to enable a global batch size of 1,024. Synchronized BN is used with group size 1.
The total training iterations is set to 100,000, and the initial learning rate is 0.001 and warms up to 0.4 during the first 10,000 iterations.
The weight decay is set to 1e-5 and momentum is set to 0.9.
Dropout with drop rate of 0.4 for the final embedding is used to prevent overfitting.

The results of two base models on the challenge test server~\cite{lfr} are shown in Tab~\ref{tab:base}.

\begin{table}[h]
\centering
\begin{center}{
\begin{tabular}{l|c c c}
\hline
 Model & Flops &Loss & TPR@FPR=1e-8  \\
 \hline
R100 & 24.22G & ArcFace & 90.972\\
PolyNet & 16.62G & ArcFace & 90.829 \\
\hline
\end{tabular}
}
\end{center}
\caption{The comparison between different base models. The Flops is computed by the public tool in \url{https://github.com/Swall0w/torchstat} (the total MAdd in the public tool). }
\label{tab:base}
\end{table}

\section{New loss function: ArcNegFace}

We introduce a new robust loss named ArcNegFace in this section. Unlike most of the recent novel losses that try to find an `optimal' logits curve to regularize the margin between embedding and class anchors, ArcNegFace takes the distance between anchors into consideration.

Define $\theta_{y_i}$ as the angle between the feature $f$ with label $y_i$ and the anchor weight $W_{y_i}$, the original ArcFace can be defined as:

\begin{eqnarray}
\begin{split}
L = -\frac{1}{n}\sum^N_{i=1}log \frac{e^{s(cos(\theta_{y_i}+m))}}{e^{s(cos(\theta_{y_i}+m))}+
\sum^n_{j=1,j\neq y_i}e^{scos\theta_j}}
\end{split}
\end{eqnarray}
where hyperparam s and m represent the scale and margin.
In order to utilize hard negative mining and weaken the influence of the error labeling,
we improve the ArcFace to ArcNegFace formulated as:
\begin{eqnarray}
\begin{split}
L = -\frac{1}{n}\sum^N_{i=1}log \frac{e^{s(cos(\theta_{y_i}+m))}}{e^{s(cos(\theta_{y_i}+m))}+
\sum^n_{j=1,j\neq y_i}e^{s(t_{j,y_i}cos\theta_j+t_{j,y_i}-1)}}
\end{split}
\end{eqnarray}
where $t_{j,y_i}$ is $\bf{G(C_j, C_{y_i})}$, $C_j$ and $C_{y_i}$ mean the cos$\theta_j$ and cos($\theta_{y_i}$+m).
The function $G(\cdot,\cdot)$ is the Gaussian function which is formulated as:

\begin{eqnarray}
\begin{split}
G(x,y) = \alpha*e^{-\frac{(x-y-\mu)^2}{2\sigma}}
\end{split}
\end{eqnarray}
where $\alpha$, $\mu$ and $\sigma$ are set to 1.2, 0 and 1, respectively.
The performance of ArcNegFace is shown in Tab~\ref{tab:loss}



\begin{table}[h]
\centering
\begin{center}{
\begin{tabular}{l|c c c}
\hline
 Model &Loss & TPR@FPR=1e-8  \\
 \hline
PolyNet & ArcFace & 90.829 \\
PolyNet& ArcNegFace & 91.639 \\
\hline
\end{tabular}
}
\end{center}
\caption{The comparison between different loss functions. }
\label{tab:loss}
\end{table}

\section{Efficient PolyFace}
Inspired by the idea of efficientnet~\cite{tan2019efficientnet}, we launch a NAS processing to expand the basic models in depth and width with the constraint of the computation budget.
Some selected results on R100 are shown in Tab~\ref{tab:search}. Note that all of the experiments are trained under the same basic setting.
Finally, we found one of the expanded PolyFace models outperforms all searched candidates with the same Flops ($\sim$28 Gflops), so we adopt it, called Efficient PolyFace, as the final backbone \footnote{Model architecture and parameters will be open-source}. Some selected results are shown in Tab~\ref{tab:cos}.

\begin{table}[t]
\centering
\begin{center}{
\begin{tabular}{l c c }
\hline
Block number & Channel number& TPR@FPR=1e-8  \\
\hline
[3,13,30,3]& [64,128,256,512]&88.652 \\
\hline
[3,13,30,3]& [72,144,288,576]&90.243 \\
\hline
[3,16,37,3]& [65,130,260,520]&90.188 \\
\hline
[3,20,46,3]& [59,118,236,472]&89.954 \\
\hline
[3,25,57,3]& [53,106,212,424]&89.875 \\
\hline
[3,13,50,3]& [61,122,244,488]&89.789\\
\hline
[3,9,19,3]& [84,168,336,672]& 89.734 \\
\hline
[3,9,31,3]& [74,148,296,592]&89.699 \\
\hline
\end{tabular}
}
\end{center}
\caption{The performance of different modified R100 models.}
\label{tab:search}
\end{table}

\begin{table}[H]
\centering
\begin{center}{
\begin{tabular}{l|c c c}
\hline
 Model &AdaBN & TPR@FPR=1e-8  \\
 \hline
Efficient PolyFace &  & 93.801 \\
Efficient PolyFace ABN & $\surd$ & \bf{94.198} \\
\hline
\end{tabular}
}
\end{center}
\caption{Performance of AdaBN. The performance 94.198 is the final submission on the leaderboard.}
\label{tab:adabn}
\end{table}

\begin{table}[h]
\centering
\begin{center}{
\begin{tabular}{l|c c c}
\hline
 Model &margin & TPR@FPR=1e-8  \\
 \hline
PolyNet & 0.5 & 90.829 \\
PolyNet & 0.3 & 91.332 \\
\hline
\end{tabular}
}
\end{center}
\caption{The performance of different margin based on ArcFace. }
\label{tab:margin}
\end{table}

\section{Bag of tricks}

\subsection{Anchor finetuning}
We introduce a new regularization term named  $anchor$ $finetuning$. Given a convergent model, we extract the features of the training set and re-init the weight $W$ in the classification layer by the mean feature of the corresponding identity. Then, the model will be finetuned based on this as shown in Tab ~\ref{tab:aug}.

\subsection{Scale \& Shift augmentations}

Data augmentation is used during the training process for all settings. The original image will be re-scaled and shifted within $\pm 1\%$ randomly.
The performance is shown in Tab~\ref{tab:aug}.

\subsection{Color jitter}
The brightness, contrast, and saturation are set to 0.125 when adding color jitter.

\subsection{Flip strategy}
The flip strategy is adopted during the training stage. During the inference stage, we extract the features for both the original and the flipped image. The final feature is the average of them. Results are shown in Tab~\ref{tab:aug}.

\subsection{Regular face}
Regular face~\cite{Zhao_2019_CVPR} is adapted to constrain the inter-class distance, but we find it can rarely bring improvement while consuming a large memory.

\subsection{Label smooth}
We explore the label smooth strategy, which is widely used in ImageNet classification. The result is shown in Tab~\ref{tab:aug}.


\begin{table*}[h]
\centering
\begin{center}{
\scalebox{0.88}{
\begin{tabular}{l|c c c c c c c|c}
\hline
 Model & ArcNegFace & Scale\&Shift aug & Flip & Regular Face~\cite{Zhao_2019_CVPR} & Label smooth& Fc finetune & Arch finetune~\cite{he2019bag} & TPR@FPR=1e-8  \\
\hline
R100 & $\surd$ &  & & & & & &81.503 \\
R100 & $\surd$ & $\surd$ & & & & & &80.59 \\
R100 & $\surd$ &  & & & & $\surd$ & &81.628 \\
R100 & $\surd$ & $\surd$ & &$\surd$ & &  & &80.819 \\
R100 & $\surd$ & $\surd$ & & &$\surd$ &  & &81.085 \\
R100 & $\surd$ & $\surd$ & & &$\surd$ &  & $\surd$&81.272 \\
R100 & $\surd$ & $\surd$ &$\surd$ & &$\surd$ &  & &81.922 \\
R100 & $\surd$ &  &$\surd$ & &$\surd$ & &$\surd$ &81.638 \\
\hline
\end{tabular}}
}
\end{center}
\caption{The comparison of the different training strategy. Note that the performance is evaluated on the old obsoleting deepglint-large test server without cleaning up the error label.}
\label{tab:aug}
\end{table*}

\subsection{AdaBN}
Considering the domain shift between the training set and the testset, we perform the AdaBN~\cite{li2016revisiting} on the convergent model to improve its performance.
Results are shown in Tab~\ref{tab:adabn}.

\subsection{Modification of margin}
We modify the margin in ArcFace and it brings a few improvements as shown in Tab~\ref{tab:margin}.

\begin{table*}[h]
\centering
\begin{center}{
\begin{tabular}{l|c c c c c c }
\hline
 Model &Flops&Blocks& Cosine decay & Stochastic depth&Color jitter& TPR@FPR=1e-8  \\
 \hline
PolyNet~\cite{zhang2017polynet} &16.62G &[10,20,10] &$\surd$& $\surd$ &$\surd$& 93.066 \\
PolyFace & 24.04G &[20,30,20] & $\surd$& $\surd$ &$\surd$& 93.729 \\
Efficient PolyFace & 28.25G &[23,38,23] & $\surd$& $\surd$ &$\surd$& 93.801 \\
\hline
\end{tabular}
}
\end{center}
\caption{The performance of cosine decay and stochastic depth based on ArcNegFace. The $Scale\&Shift$ $aug$ and $Flip$ are adopted in these experiments.
ArcNegFace with margin 0.3 is used.}
\label{tab:cos}
\end{table*}
\subsection{Cosine learning rate and stochastic depth}
\begin{figure}
\centering
\includegraphics[width=1\linewidth]{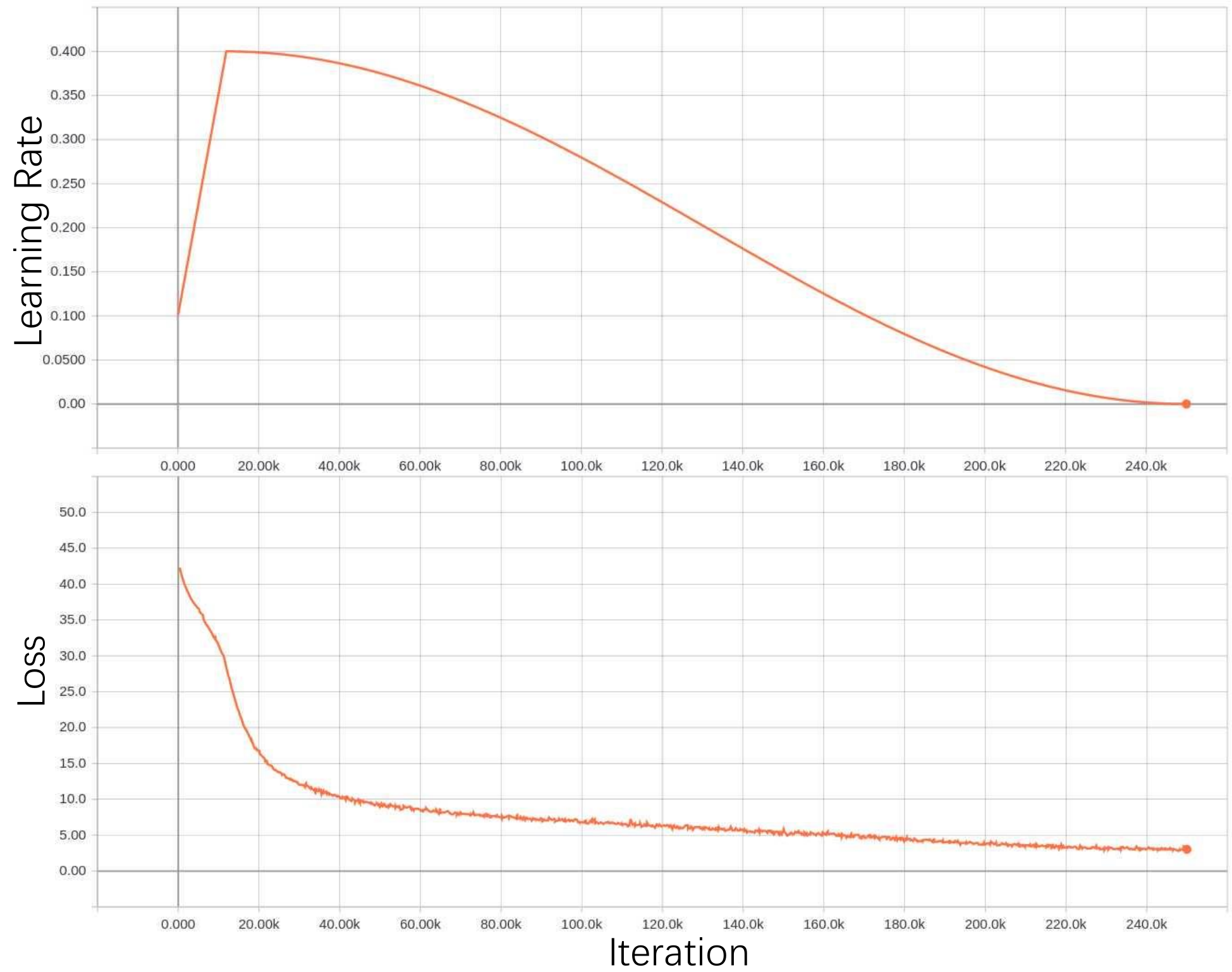}
   \caption{The details of cosine decay.}
\label{fig:cosdecay}
\end{figure}

We explore the cosine learning rate decay and stochastic depth~\cite{huang2016deep} to achieve further gain. The keep rate in stochastic depth is set to 0.8 in all experiments.
The function of learning rate \textit{w.r.t.} iteration is shown in Fig~\ref{fig:cosdecay}, and results are shown in Tab~\ref{tab:cos}.
The losses during the training of basic PolyFace is shown in Fig~\ref{fig:cosdecay}.

\section{Enhanced quality aware network for video face recognition}
To generate the robust video representation for set-to-set recognition in IQIYI track~\cite{lfr}, inspired by QAN and RQEN~\cite{Liu_2017_CVPR,song2018region}, we propose a new quality estimation strategy called enhanced quality aware network (QAN++) to approximate the quality of each image. The representation of the image set can be aggregated by the weighted sum of frame representations with the assistant of the image quality.

Different from the subjective quality judgment of image, our method assigns the image quality from the characteristics of feature discrimination.
Define the dataset $D$ with $C$ identities and the weight anchor $W_i$, $i\in[1,C]$ in the final classification layer, the quality of image $I$ with ID $c$ can be computed by:
\begin{eqnarray}
\begin{split}
Q_{I} = \frac{cos(F_I,W_c)}{max\{cos(F_I,W_j)|j\in[1,C],j\neq c\}}
\end{split}
\end{eqnarray}
The image quality is computed on the training set and in order to obtain the image quality during the inference stage, we add a lightweight quality generation branch to regress the quality value computed on the training set.
To better regress the quality, we normalize it as:
\begin{eqnarray}
\begin{split}
Q_{I} = \sigma(\frac{Q_{I}-mean(Q)}{std(Q)})
\end{split}
\end{eqnarray}
where $\sigma$($\cdot$), $mean(Q)$ and $std(Q)$ mean the sigmoid function, mean value and standard deviation value in the whole training set respectively.
The L2 loss is adopted as the training loss.

During the inference stage, given the video $I_i,i\in[1,n]$ where n means the total image number and the corresponding feature representation $F_i$, we extract the quality value $Q_i$ of $I_i$.
The quality value will be re-scaled by:
\begin{eqnarray}
\begin{split}
Q_{i} = K\cdot Q_i+B
\end{split}
\end{eqnarray}
\begin{eqnarray}
\begin{split}
K = \frac{1}{max\{Q_i\}-min\{Q_i\}}, i\in[1,n]
\end{split}
\end{eqnarray}
\begin{eqnarray}
\begin{split}
B = 1- K\cdot max\{Q_i\}, i\in[1,n]
\end{split}
\end{eqnarray}
Finally, the video-level feature can be aggregated by:
\begin{eqnarray}
\begin{split}
F = \sum^n_i \frac{Q_{i}\cdot F_i}{Q_{i}}, i\in[1,n] \label{1}
\end{split}
\end{eqnarray}
If the image number n in the image set is less than 3, we directly adopt Eq~\ref{1} to aggregate them without re-scaling the quality value. 

\subsection{Performance of different aggregation strategies}
We evaluate the effectiveness of the proposed quality estimation strategy on IQIYI in LFR.
Results are shown in Tab~\ref{tab:video}.
We embed a new quality branch into PolyFace. The new branch looks like a tiny version of ResNet-18. The block number in each stage is $[2,2,2,2]$ and
the channel number in each stage is set to [8,16,32,48]. 
We add a fully connected layer with output number 1 after the global average pooling to regress the quality.
The flops of the quality net is $81.9$ Mflops and the input is the same as the PolyFace.

\begin{table}[H]
\centering
\begin{center}
\scalebox{1}{
\begin{tabular}{l|c c| c }
\hline
Model (w/o ABN)& Deepglint & aggregation & IQIYI \\
\hline
R100 & 92.433&Avg & 65.843 \\
R100 & 92.433&Weighted Sum & 67.381 \\
R100 & 92.433&Top1 Quality & 65.217 \\
R100 & 92.433&QAN++ & 69.048 \\
PolyFace &93.729 & QAN++ & \bf{72.981}\\
\hline
\end{tabular}}
\end{center}
\caption{Comparison with different quality strategies on IQIYI-large track in LFR. The performance 72.981 won the \textbf{1st place} in this competition.}
\label{tab:video}
\end{table}

\section{Conclusion}
In this article, we show the details of our solution to the ICCV19-LRF challenge.
For the image-based and video-based tracks, We introduce a new backbone Efficient PolyFace and a new loss function ArcNegFace.
For the video based track, we propose a novel quality estimator QAN++ to generate quality score for each frame. 
Besides, we also explore some useful tricks in face recognition model.
Results on the challenge test server demonstrate the effectiveness of the proposed methods.


{\small
\bibliographystyle{ieee}
\bibliography{egbib}
}

\end{document}